\documentclass[conference]{IEEEtran}
\IEEEoverridecommandlockouts
\usepackage{cite}
\usepackage{amsmath,amssymb,amsfonts}
\usepackage{algorithmic}
\usepackage{algorithm}
\usepackage{graphicx}
\usepackage{textcomp}
\usepackage{xcolor}
\usepackage{makecell}
\usepackage{pifont}
\usepackage{float}
\usepackage{caption}

\def\BibTeX{{\rm B\kern-.05em{\sc i\kern-.025em b}\kern-.08em
    T\kern-.1667em\lower.7ex\hbox{E}\kern-.125emX}}

\begin{document}

\title{SpatialMe: Stereo Video Conversion Using Depth-Warping and Blend-Inpainting}


\author{
    \IEEEauthorblockN{Jiale Zhang$^{1}$, Qianxi Jia$^{1}$, Yang Liu$^1$, Wei Zhang$^1$, Wei Wei$^{1*}$, Xin Tian$^1$}
    \IEEEauthorblockA{$^1$ JD.com, Beijing, China}
    \IEEEauthorblockA{\{zhangjiale90, jiaqianxi1, liuyang1605, zhangwei1282, weiwei20, tianxin84\}@jd.com}
}


\twocolumn[{
\renewcommand\twocolumn[1][]{#1}
\maketitle
\begin{center}
    \captionsetup{type=figure}
    \includegraphics[width=\linewidth]{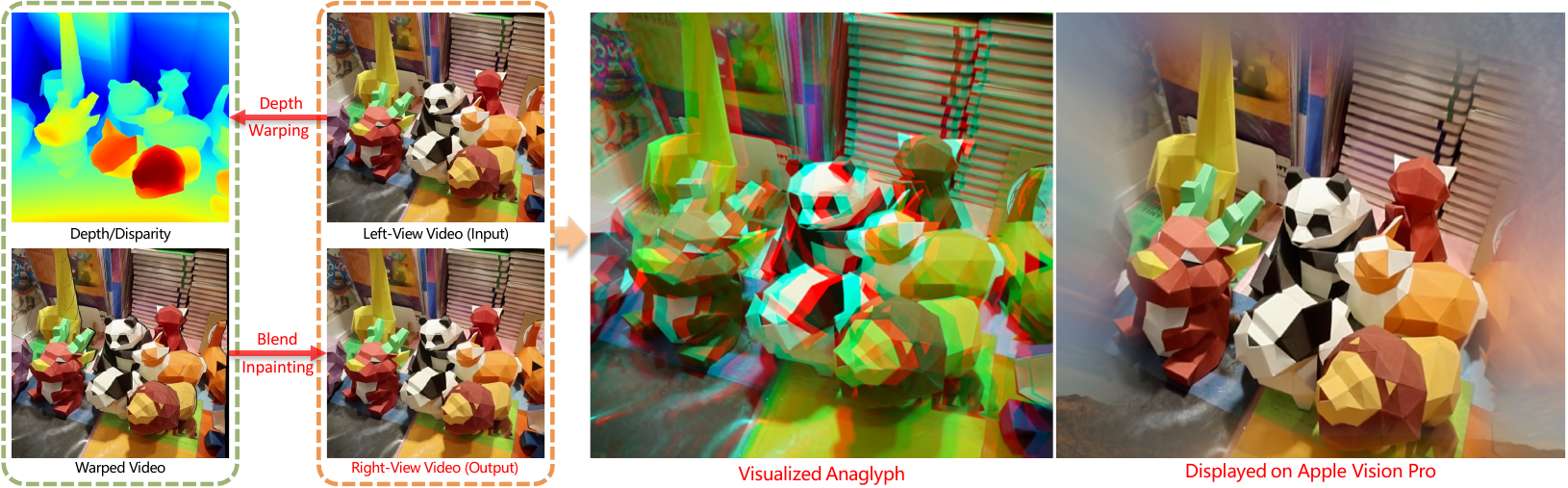}
    \captionof{figure}{Our method performs stereo conversion for any 2D video to stereo format via depth-warping and blend-inpainting, enabling viewing on 3D glasses (anaglyph) or AR/VR devices like Apple Vision Pro.}
    \label{fig_figure1}
\end{center}}]

\begin{abstract}
Stereo video conversion aims to transform monocular videos into immersive stereo format. Despite the advancements in novel view synthesis, it still remains two major challenges: i) difficulty of achieving high-fidelity and stable results, and ii) insufficiency of high-quality stereo video data. In this paper, we introduce SpatialMe, a novel stereo video conversion framework based on depth-warping and blend-inpainting. Specifically, we propose a mask-based hierarchy feature update (MHFU) refiner, which integrate and refine the outputs from designed multi-branch inpainting module, using feature update unit (FUU) and mask mechanism. We also propose a disparity expansion strategy to address the problem of foreground bleeding. Furthermore, we conduct a high-quality real-world stereo video dataset---StereoV1K, to alleviate the data shortage. It contains $1000$ stereo videos captured in real-world at a resolution of $1180\times1180$, covering various indoor and outdoor scenes. Extensive experiments demonstrate the superiority of our approach in generating stereo videos over state-of-the-art methods.

\end{abstract}


\section{Introduction}
\label{sec:intro}
With the rapid advancement of AR/VR devices, the demand for stereo videos with immersive experiences is on the rise. Converting existing 2D videos to immersive 3D videos is a way to supply content rapidly. As an inherently ill-posed problem, stereo conversion has undergone significant evolution with the advent of deep learning, shifting from early convolutional neural network methodologies to intricate diffusion designs. This process primarily involves generating the right view from the left view and compensating for occluded information. 
Existing conversion methods mainly rely on monocular depth estimation~\cite{ranftl2020towards,ranftl2021vision,yang2024depth1,yang2024depth2} combined with inpainting~\cite{liu2018image,zhou2023propainter} or diffusion model~\cite{zhao2024stereocrafter,wang2024stereodiffusion,lv2024spatialdreamer} to achieve the generation of the right view. However, they often lead to heavy artifacts or uncontrolled results.

In this paper, we present a novel framework for stereo video conversion. Specifically, at the depth-warping stage, we first enhance the depth estimation model from \cite{yang2024depth2} to generate accurate and stable depth/disparity maps, which guide the input video to generate warped one and corresponding occlusion masks. During the blend-inpainting stage, we introduce a disparity expansion strategy to address foreground bleeding by providing background pixels at edge regions. Unlike previous single-branch approaches, we propose a multi-branch inpainting module incorporates traditional, deep learning, and disparity expansion strategies. To leverage the strengths of each branch, we further design a MHFU refiner with feature update units to integrate and optimize these outputs, obtaining the final result. As illustrated in Fig.~\ref{fig_figure1}, our model achieves high-fidelity stereo conversion suitable for viewing with 3D glasses or AR/VR devices like Apple Vision Pro.

\begin{figure*}[!t]
    \centering
    \includegraphics[width=\linewidth]{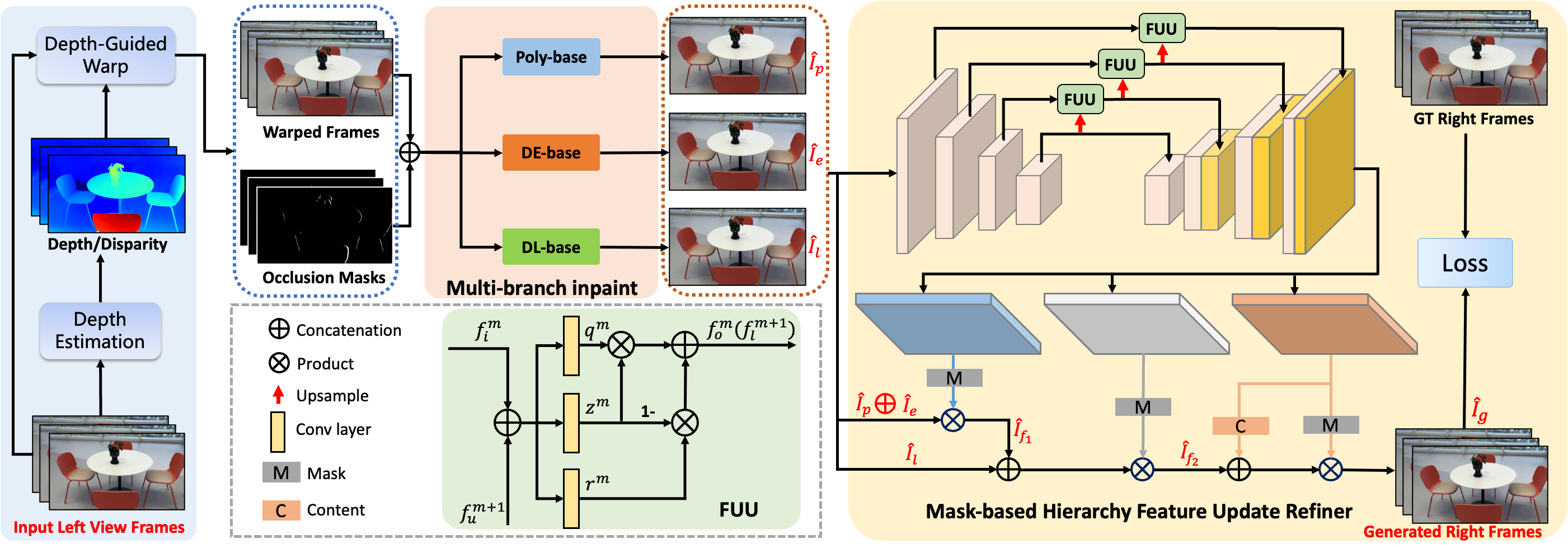}
    \caption{Overview of our framework, which consists of two stages. First, a depth estimation model predicts the depth of the left-view frames, guiding the warping of right-view frames and generating occlusion masks. In the second stage, a multi-branch inpainting module, composing traditional (Poly-based), deep learning (DL-based), and disparity expansion (DE-based) branches, fills the occluded regions based on the occlusion masks. The inpainted results are then fused and refined via a mask-based hierarchical feature update refiner, generating the final right-view frames.}
    \vspace{-10pt}
    \label{fig_architecture}
\end{figure*}

Another challenge of this task is the lack of high-quality 
real-world stereo video data. To address it, we conduct a large-scale benchmark stereo video dataset captured in various indoor and outdoor scenes, using a professional spatial video camera.
To sum up, the contributions of this work include:

\begin{itemize}
\item We present a novel mask-based hierarchical feature update refiner that works with our multi-branch inpaint module to achieve high-fidelity stereo conversion.
\item We propose a disparity expansion strategy to improve inpainting accuracy by providing background information to foreground edge regions.
\item We introduce StereoV1K, a high-quality, large-scale stereo video dataset with $1000$ videos captured in various real-world scenes, featuring over $500,000$ frames at a resolution of $1180\times1180$.
\end{itemize}

\section{Related Work}
\label{sec:intro}
\subsection{Depth Estimation and Inpainting}
Converting monocular videos to stereo has been a vibrant research area in computer vision and graphics over the past decade. With advancements in deep learning, many methods now leverage depth estimation and inpainting to achieve this conversion. Shih et al.~\cite{shih20203d} introduce a method of layered depth image (LDI)~\cite{shade1998layered} to select depth edges for inpainting. 
Jampani et al.~\cite{jampani2021slide} and Wang et al.~\cite{wang20223d} also use LDI representation with fewer layers to generate novel views. Han et al.~\cite{han2022single} and Pu et al.~\cite{pu2023sinmpi} use depth estimation and multiplane images (MPI)~\cite{zhou2018stereo} representation for novel view synthesis. However, these approaches rely on naive inpainting techniques, which fall short when dealing with complex scenes.

\subsection{Diffusion Models}
Recently, the advancement of diffusion models has spurred significant progress in the stereo video domain. Lv et al.~\cite{lv2024spatialdreamer} introduces a depth-driven video generation module using diffusion models, employing a forward-backward rendering mechanism to produce stereo videos. Wang et al.~\cite{wang2024stereodiffusion} use stereo pixel shifting operations to extract latent vectors for generating the right view from the left with diffusion models. Zhao et al.~\cite{shi2024stereocrafter} and Shi et al.~\cite{zhao2024stereocrafter} leverages video diffusion priors to optimize temporal consistency and view alignment in stereo video generation. Despite these advancements, challenges such as uncontrollable image generation and high computational demands remain to be addressed.

\section{Proposed Method}
    \subsection{Overview}
    In this paper, we aim to design an architecture that converts a sequence of frames from the left view into the corresponding right view, thereby creating a stereo frame sequence, which can subsequently be used for display on VR/AR devices or 3D glasses. To this end, we introduce our proposed framework, as shown in Fig.~\ref{fig_architecture}. At stage one, a depth estimation module is used to obtain the depth/disparity, which guides the warp of input frames to generate warped frames and corresponding masks.
    In order to improve both the generalization capability and accuracy of depth estimation for video sequences, we enhance it by fine-tuning the approach outlined in\cite{yang2024depth2}. At stage two, a multi-branch inpainting module processes the warped sequence and occlusion masks to produce three inpainting results. To leverage the strengths of each inpainting output, we introduce a mask-based hierarchical feature update refiner, along with a specially designed feature update unit, to integrate these outputs, yielding the final refined result.

    \vspace{-1pt}\subsection{Multi-Branch Inpaint Module}
    We propose a multi-branch inpainting module that combines three distinct inpainting strategies: polygon interpolation-based (Poly-based), deep learning-based (DL-based), and disparity expansion-based (DE-based) inpainting. Our motivation stems from the observation that each strategy has unique strengths and weaknesses, but individually, they often lead to unsatisfactory results. By integrating these complementary branches, we achieve superior inpainting performance.

    \vspace{2pt}\noindent\textbf{Poly-Based Inpaint.}
    In practice, traditional interpolation-based inpainting methods consistently generate stable content across frame sequences by using deterministic formulas that derive inpainted pixels from surrounding values. This stability makes them effective for inpainting small areas, preserving text clarity, and enhancing video stability by reducing shake. Our polygon-based inpainting branch utilizes an advanced polygon interpolation method \cite{bobthirygithub}, which outperforms naive interpolation techniques. It treats image rows as polylines, adjusting disparity to morph these polylines before rasterization. However, while these methods excel in stability, they struggle with generating fine details at object edges or over large areas, often resulting in stretched artifacts.

    \vspace{2pt}\noindent\textbf{DL-Based Inpaint.}
    To address the limitations of the polygon-based inpainting branch, we incorporate a deep learning-based inpainting branch, which excels at generating reasonable content for areas requiring filling. To ensure stability and consistency across frame sequences, we adopt a flow-based video inpainting approach, ProPainter \cite{zhou2023propainter}, which leverages content from adjacent frames for more coherent and natural results. However, while deep learning approaches excel at generating realistic content, they often extend foreground elements into foreground-background edge regions, disrupting the spatial structure. Ideally, these regions should be filled with background content. This issue arises because these methods generate pixels based on surrounding context, leading to errors at the edges where background information is sparse.

    \vspace{2pt}\noindent\textbf{DE-Based Inpaint.}
    To overcome the issue of foreground elements encroaching into foreground-background edge regions during inpainting, we introduce a disparity expansion-based inpainting branch. This branch focuses on enhancing the representation of critical boundary areas by extending the disparity map to incorporate background information, thereby improving inpainting accuracy.

\begin{figure}[!t]
    \vspace{-7pt}
        \centering
        \includegraphics[width=3.35in, angle=0]{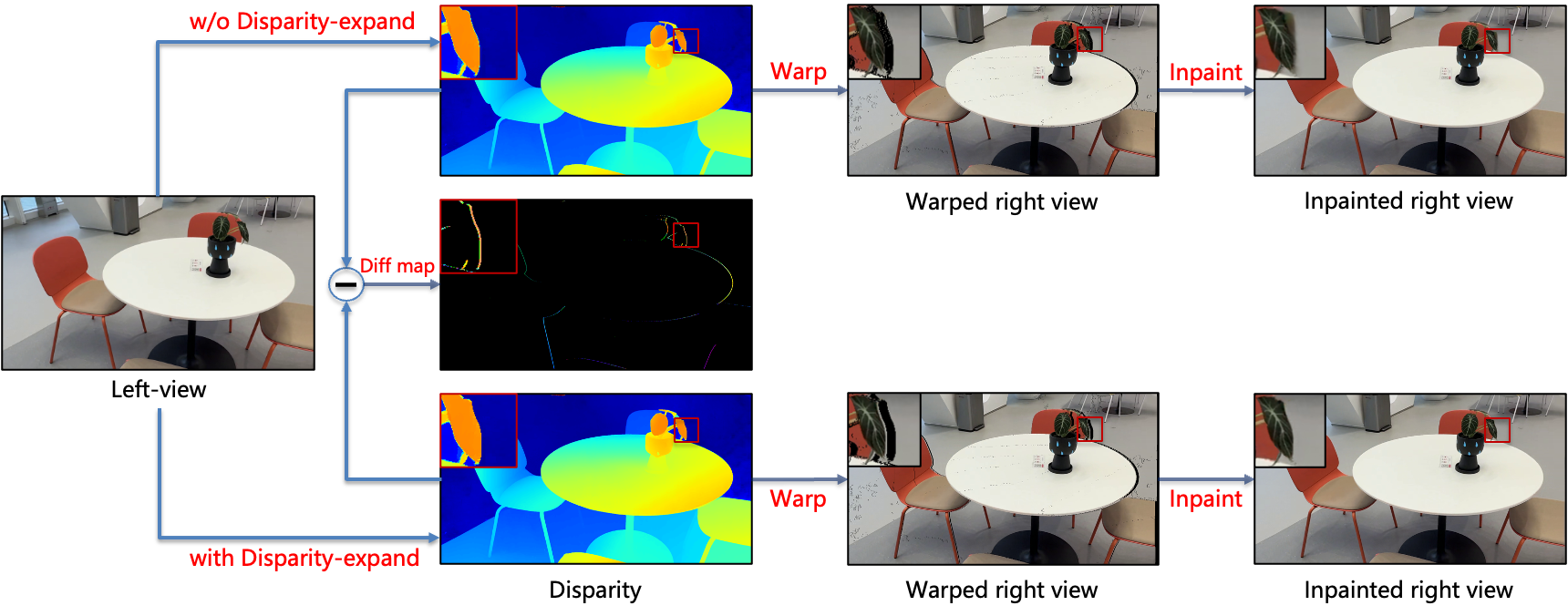}
        \caption{Illustration of the disparity expansion strategy. The result of using this strategy ensures more realistic inpainting by accurately using the background context}
        \vspace{-10pt}
        \label{fig_DE}
    \end{figure}

    As outlined in Algorithm \ref{alg:DE}, the disparity expansion strategy first identifies regions where the foreground and background intersect by analyzing the disparity map. It then extends the edges of the disparity map to include more background details at the warped foreground edges. This ensures access to richer background context during inpainting, maintaining the spatial integrity of the scene. By reducing the propagation of foreground elements into areas meant for background content, this method preserves the visual structure and coherence of the frame sequence. Fig.~\ref{fig_DE} demonstrates the improved results with the disparity expansion strategy, showing enhanced background detail along warped foreground edges and significantly reducing foreground intrusion. However, in occluded areas with complex depth, the expanded background elements may be incorrect, leading to inaccuracies in the inpainted content.

    \begin{algorithm}[!h]
        \caption{Disparity Expansion}
        \label{alg:DE}
        \renewcommand{\algorithmicrequire}{\textbf{Input:}}
        \renewcommand{\algorithmicensure}{\textbf{Output:}}
        
        \begin{algorithmic}[1]
            \REQUIRE Disparity ${I}$, Radius $k$, Threshold $\lambda$  
            \ENSURE Expanded Disparity $\hat{I}$    

            \STATE $E = \text{Canny}(I)$, $\hat{I} = \text{deepcopy}(I)$
            \STATE $row, col = I.\text{shape}[0], I.\text{shape}[1]$
            
            \FOR{$i, j \in \text{np.argwhere}(E \neq 0)$ \AND $k \leq j < (col - k)$}
                \IF{$I_{i,j-1} - I_{i,j+1} > \lambda$ \AND $k \leq i < (row - k)$}
                    \STATE $\hat{I}_{(i-k:i+k),(j-k:j+k)} = I_{i,j-1}$
                \ENDIF
            \ENDFOR
            
            \RETURN $\hat{I}$
        \end{algorithmic}
    \end{algorithm}

    \subsection{Mask-Based Hierarchy Feature Update Refiner}
    As mentioned above, each branch in our multi-branch module has its own strengths and weaknesses. To enhance the overall generation results, we introduce a novel mask-based hierarchical feature update refiner, detailed in Fig.~\ref{fig_architecture}. The refiner takes three inputs ($\hat{I}_{p}$, $\hat{I}_{e}$, and $\hat{I}_{l}$) from the multi-branch inpainting module and processes them through a series of three convolutional layers to output distinct single-channel attention masks and RGB content. The final output is synthesized by fusing $\hat{I}_{p}$, $\hat{I}_{e}$, and $\hat{I}_{l}$ with these masks and content.

    This process is detailed as follows: Let ${M}_{1}$, ${M}_{2}$, and ${M}_{3}$ denote the first, second, and third mask generated by the refiner, ${C}$ denote the generated RGB content. Then we obtain the fusion result $\hat{I}_{g}$ by
    \vspace{-2pt}
    \begin{equation}
        \hat{I}_{f_{1}} = M_1 \cdot \hat{I}_p + \left( 1 - M_1 \right) \cdot \hat{I}_e\,,\label{f1}
    \vspace{-2pt}
    \end{equation}
    \begin{equation}
        \hat{I}_{f_{2}} = M_2 \cdot \hat{I}_{f_{1}} + \left( 1 - M_2 \right) \cdot \hat{I}_l\,,\label{f2}
    \vspace{-2pt}
    \end{equation}
    \begin{equation}
        \hat{I}_{g} = M_3 \cdot \hat{I}_{f_{2}} + \left( 1 - M_3 \right) \cdot C\,,\label{f3}
    \vspace{-2pt}
    \end{equation}
    where $\cdot$ denotes element-wise multiplication. The generated masks $M$ effectively guide the network to focus precisely on foreground-background boundaries. By linearly fusing advantageous content from different inputs in these critical regions, the refiner achieves better results. Simultaneously, the content image $C$ flexibly integrates existing information while accurately targeting edge inpainted areas that require further enhancement through new content generation. Consequently, our mask-based hierarchical feature update refiner generates results that are both natural and realistic.

    \vspace{2pt}\noindent\textbf{Feature Update Unit.}
    High-resolution features capture detailed local information, while low-resolution features provide abstract global structures. To effectively fuse these multi-resolution representations, we introduce Feature Update Unit (FUU), which performs layer-by-layer fusion and updates encoder features across resolutions. The updated features are then skip-connected to corresponding decoder layers, resulting in fully integrated and enhanced feature representations.

    To illustrate the process, we take the $m$-th encoder layer as an example. Let ${f^m}_i$ and ${f^m}_o$ denote the input encoder feature and output feature of FUU in $m$-th layer. We first compute the upsampled encoder feature of $m+1$-th layer ${f^{m+1}}_u$, then these features are integrated and updated by
    \vspace{-2pt}
    \begin{equation}
        {z^m} = \sigma \left(Conv \left( \left[ {f^m}_{i}, {f^{m+1}}_{u} \right] \right) \right)\,,
    \vspace{-2pt}
    \end{equation}
    \begin{equation}
        {q^m} = \tanh \left(Conv \left( \left[ {f^m}_{i}, {f^{m+1}}_{u} \right] \right)\right) \,,
    \vspace{-2pt}
    \end{equation}
    \begin{equation}
        {r^m} = \tanh \left(Conv \left( \left[ {f^m}_{i}, {f^{m+1}}_{u} \right] \right)\right) \,,
    \vspace{-2pt}
    \end{equation}
    \begin{equation}
        {{f^m}_{o}} = q^m \cdot z^m + \left( 1 - z^m \right) \cdot r^m\,,
    \vspace{-2pt}
    \end{equation}
    where $\cdot$ denotes element-wise multiplication, $Conv$ denotes the convolution layer, and $\sigma$ denotes the sigmoid function.
    The output of FUU in the $m$-th layer is denoted as ${f^m}_{o}$. Within this framework, ${q^m}$ and ${r^m}$ represent the updated high and low-resolution features. The fusion process is guided by the mask ${z^m}$, which dynamically adjusts the weights, enabling multi-level features to be iteratively updated and fused, thereby enhancing the accuracy and realism of generations.
    
    \vspace{-2pt}\subsection{Loss Function}
    Three loss terms are defined to optimize our model, including the content, perceptual, and adversarial loss.

    We use $\ell_1$ loss to represent the content loss. To improve the quality of inpainted areas and mitigate mismatches elsewhere, we employ the occlusion mask ($O$) to apply separate weights to known areas and occlusions.
    It is defined by
    \begin{equation}
    \vspace{-2pt}
        \mathcal{L}_{c} = \alpha \sum_{x \in O} \left\| {I}_{g}(x) - \hat{I}_{g}(x) \right\|_{1} \ + \sum_{x \notin O} \left\| {I}_{g}(x) - \hat{I}_{g}(x) \right\|_{1} \,,
    \vspace{-2pt}
    \end{equation}
    where $\hat{I}_{g}$ is the final result and ${I}_{g}$ is the ground truth, $\alpha = 10$. 

    Optimizing the model using only $\ell_1$ loss can lead to blurred outputs. To address this, we incorporate perceptual loss \cite{johnson2016perceptual} as an additional constraint. We input the generated results and their corresponding ground truth into a pre-trained VGG19 model \cite{simonyan2014very}, extracting features from the $conv2\_1$, $conv3\_1$, and $conv4\_1$ layers. The perceptual loss is then computed as:
    \vspace{-3pt}
    \begin{equation}
        \mathcal{L}_{per} = \sum_{j} \left\| \phi_j \left({I}_{g}\right) - \phi_j \left(\hat{I}_{g}\right) \right\|_{1} \,.
    \vspace{-3pt}
    \end{equation}
    where $\phi_j \left(\cdot\right)$ denotes the features of $VGG19$ in the $j$-th layer.

    We also adopt the adversarial loss \cite{goodfellow2020generative} to further improve the realism of results, it is obtained by
    \vspace{-2pt}
    \begin{equation}
        \mathcal{L}_{adv} =  E_{{\hat{I}_{g}}\sim P_g}\left[ D\left(\hat{I}_{g}\right) \right] - E_{{{I}_{g}}\sim P_{data}}\left[ D\left({I}_{g}\right) \right]\,,
    \vspace{-1pt}
    \end{equation}
    In the adversarial process, the generator aims to produce realistic fake samples to maximize the objective function, while the discriminator $D$ seeks to accurately differentiate between real and fake samples, minimizing the objective function.

    Our total loss is then defined as
    \vspace{-3pt}
    \begin{equation}
        \mathcal{L}_{total} = \lambda_{1} \cdot \mathcal{L}_{c} + \lambda_{2} \cdot \mathcal{L}_{per} + \lambda_{3} \cdot \mathcal{L}_{adv}\,,
    \vspace{-3pt}
    \end{equation}
    where we prioritize ${L}_{c}$ and ${L}_{per}$ as the main objective, supplemented by ${L}_{adv}$.
    Empirical evaluation shows that $\lambda_{1}=10$, 
    $\lambda_{2}=2$, and $\lambda_{3}=0.1$ can provide an optimal balance.

    \begin{table}
        {\caption{{Comparison between our StereoV1K and other related datasets.
        }}
        \label{tab:dataset}
        \centering
        \begin{tabular}{l@{\hskip 0.4cm}c@{\hskip 0.4cm}c@{\hskip 0.4cm}c@{\hskip 0.45cm}c}
            \hline
            {Dataset} & \makecell{Frames} & \makecell{Resolution} & \makecell{Available} &  \makecell{Data Source}\\ \hline
            \hline
            KITTI\cite{menze2015object}   & 400 & 1242$\times$375 &\checkmark & Auto Driving \\	
            Sintel\cite{butler2012naturalistic}  & 1064 &1024$\times$436 &\checkmark  & Synthetic \\
            3D Movies\cite{ranftl2020towards} & 75K &1920$\times$1080  & \ding{55} & 3D Movies  \\
            SceneFlow\cite{mayer2016large} & 39K  &960$\times$540 &\checkmark & Synthetic  \\
            StereoV1K(Ours)  &\bf{500K}  &\bf{1180$\times$1180} &\bf{\checkmark} & \bf{Real-World}  \\	
            \hline		
        \end{tabular}}
        \vspace{-10pt}
    \end{table}

    \begin{figure}[!t]
        \centering
        \includegraphics[width=3.35in, angle=0]{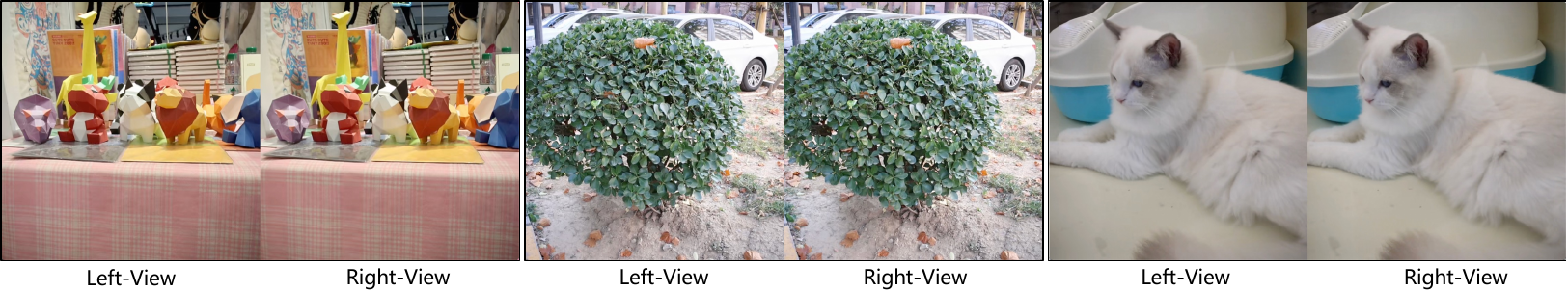}
        \caption{Illustration of the side-by-side videos in StereoV1K.}
        \label{fig_JD3DV}
        \vspace{-10pt}
    \end{figure}
    
    \section{StereoV1K Dataset}
    
    
    To address the lack of real-world video-based stereo datasets, we introduce StereoV1K, a large-scale benchmark stereo video dataset captured in real-world including various scenes. To the best of out knowledge, this is the first high-quality real-world stereo video datasets.

    For StereoV1K, we use the Canon RF-S$7.8mm$ F$4$ STM DUAL lens \footnote{\vspace{-15pt} https://www.canon-europe.com/lenses/rf-s-7-8mm-f4-stm/} to capture spatial videos in both indoor and outdoor scenes, as illustrated in Fig.~\ref{fig_JD3DV}. The dual lenses, with identical parameters, ensure high-quality stereo videos with consistent resolution, clarity, and tone across the left and right views. We also tested the iPhone series, {\it e.g.}, iPhone $16$ Pro, but found its dual cameras unsuitable due to variations in resolution, clarity, and tone with different parameters.
    
    StereoV1K contains $1000$ 3D side-by-side videos featuring various indoor and outdoor scenes. Each video is $1180\times1180$ in resolution, approximately $20$ seconds long, and recorded at $50$ fps, resulting in a selection of over $500,000$ frames. From these, $955$ videos are selected for training and $45$ for testing.

    In Table~\ref{tab:dataset}, we present a comparison between our StereoV1K dataset and other related datasets, such as KITTI \cite{menze2015object}, Sintel \cite{butler2012naturalistic}, 3D Movies\cite{ranftl2020towards}, and SceneFlow \cite{mayer2016large}. Unlike these datasets, which source their data from movies, software, or are restricted to driving scenes, StereoV1K offers over $500$K high-resolution frames captured entirely in real-world environments across diverse scenes. This significantly enhances the versatility and applicability of dataset for various research and development.
    
    To establish the correspondence between the left and right views in each side-by-side video, we employ the stereo matching method IGEV \cite{xu2023iterative}, which generates disparity maps that guide the warping of input frames.

    \section{Results and Discussions}

    \subsection{Implementation Details}
    Our model is trained on the StereoV1K using the Adam optimizer \cite{kingma2014adam}. Training is initiated from scratch with a constant learning rate of $0.0001$, and hyper-parameters $\beta_{1} = 0.5$, $\beta_{2} = 0.999$. The network is trained with a batch size of $4$ over $300K$ iterations on a single NVIDIA A$100$ GPU, completing in approximately three days.
    We use four metrics for evaluation: Mean Absolute Error (MAE), Learned Perceptual Image Patch Similarity (LPIPS), Peak Signal-to-Noise Ratio (PSNR), and Structural Similarity Index Measure (SSIM). 
    
    \subsection{Quantitative Comparison}

    \begin{table}
        \caption{Quantitative comparison with several state-of-the-art methods. The best results are in \bf{bold}.
        }
        \label{tab:comparison}
        \centering
        \begin{tabular}{l@{\hskip 0.6cm}c@{\hskip 0.5cm}c@{\hskip 0.5cm}c@{\hskip 0.5cm}c}
            \hline
            {Method} & {MAE $\downarrow$} & {LPIPS $\downarrow$} & {SSIM $\uparrow$} & {PSNR $\uparrow$}\\ \hline
            \hline
            3D-Photography\cite{shih20203d}  & 0.0379 & 0.1121 & 0.8163 & 22.228\\	
            Stereo Diffusion\cite{wang2024stereodiffusion}   & 0.0436 & 0.1501 & 0.7702  & 23.019\\
            AdaMPI\cite{han2022single}  & 0.0342 & 0.0608 & 0.7911  & 24.254\\
            Ours  &\bf{0.0318} &\bf{0.0478} &\bf{0.8522}  &\bf{31.445}\\	
            \hline		
        \end{tabular}
        \vspace{-10pt}
    \end{table}

    \begin{figure}[!t]
        \centering
        \includegraphics[width=3.35in, angle=0]{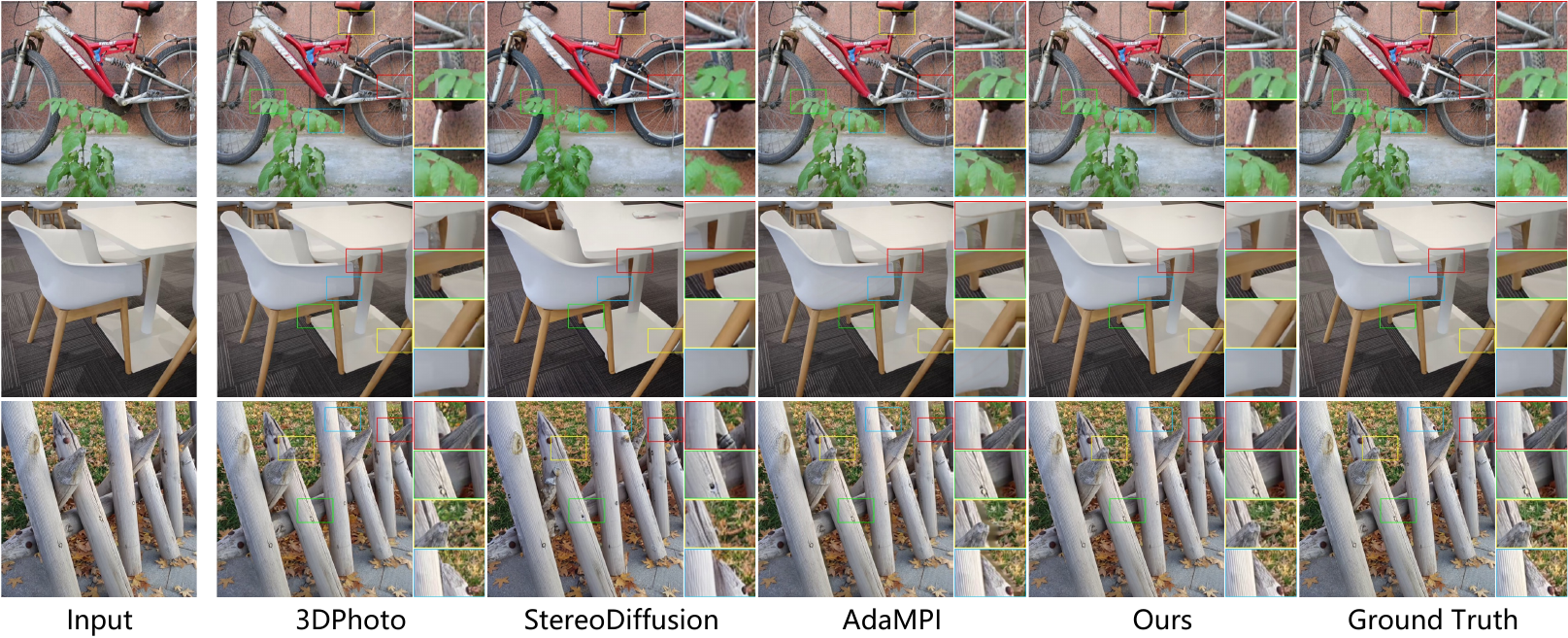}
        \caption{Qualitative comparison with several state-of-the-art methods. Our approach obtains significantly better results. }
        \vspace{-10pt}
        \label{fig_compare}
    \end{figure}
    
    There are few approaches available for direct comparison with ours, we thus select four state-of-the-art methods for comparison: 3D-Photography\cite{shih20203d}, AdaMPI\cite{han2022single}, and Stereo Diffusion\cite{wang2024stereodiffusion}, using their official implementations. We conduct evaluations on the StereoV1K dataset. Given that our test set contains $45$ videos with a total of approximately $45,000$ frames, and due to the high runtime of these methods, we perform evaluations with every $20$th frame in each video. 
    Quantitative results are presented in  Table~\ref{tab:comparison}, where lower MAE and LPIPS scores, along with higher PSNR and SSIM scores, indicate better generations. Our approach achieves outstanding performance across all metrics, outperforming other methods by a large margin, consistent with our qualitative results. This demonstrates the effectiveness of our multi-branch fusion and refinement framework with disparity expansion strategy, and the high quality of our real-world dataset, StereoV1K.

    \subsection{Qualitative Comparison}
    We also provide the results of qualitative comparison in Fig.~\ref{fig_compare}. The grid representation of 3D-Photography results in overly sharp edges and image distortion, as well as some mistake generations. AdaMPI suffers from severe edge stretching and noticeable artifacts. StereoDiffusion transforms the input image into latent features and warps them to generate the right-view image. It relies heavily on inversion performance, which leads to instability and inconsistency in the generated content. Instead, our approach generates more realistic, accurate, and natural results with reduced blur and fewer artifacts.
    
    \subsection{Ablation Study}
    To evaluate the effectiveness of each module in our method, we conducted an ablation study. The variants we compared mainly include the following:
    \begin{itemize}
        \item Baseline-Poly. We only use the poly-based interpolation.
        
        
        \item Baseline-DL. We use only the deep learning-based model.
        
        \item Baseline-DL + DE. We use Propainter with the proposed disparity expansion strategy.
        
        \item Full-Model. Our full model is used, including the mask-based hierarchy feature update refiner (MHFU), disparity expansion strategy, and multi-branch inpainting module.
    \end{itemize}

    \begin{table}
        \caption{Evaluation results of our ablation study.
        }
        \label{tab:ablation}
        \centering
        \begin{tabular}{l@{\hskip 0.6cm}c@{\hskip 0.6cm}c@{\hskip 0.6cm}c@{\hskip 0.6cm}c}
            \hline
            {Method} & {MAE $\downarrow$} & {LPIPS $\downarrow$} & {SSIM $\uparrow$} & {PSNR $\uparrow$}\\ \hline
            \hline
            Baseline-Poly   & 0.0381 & 0.0701 & 0.8249 & 29.488\\	
            Baseline-DL  & 0.0377 & 0.0683 & 0.8260  & 29.576\\
            Baseline-DL + DE   & 0.0371 & 0.0668 & 0.8284  & 29.896\\
            Full-Model  &\bf{0.0320} &\bf{0.0481} &\bf{0.8507}  &\bf{31.412}\\	
            \hline		
        \end{tabular}
        \vspace{-10pt}
    \end{table}

    \begin{figure}[!t]
        \centering
        \includegraphics[width=3.35in, angle=0]{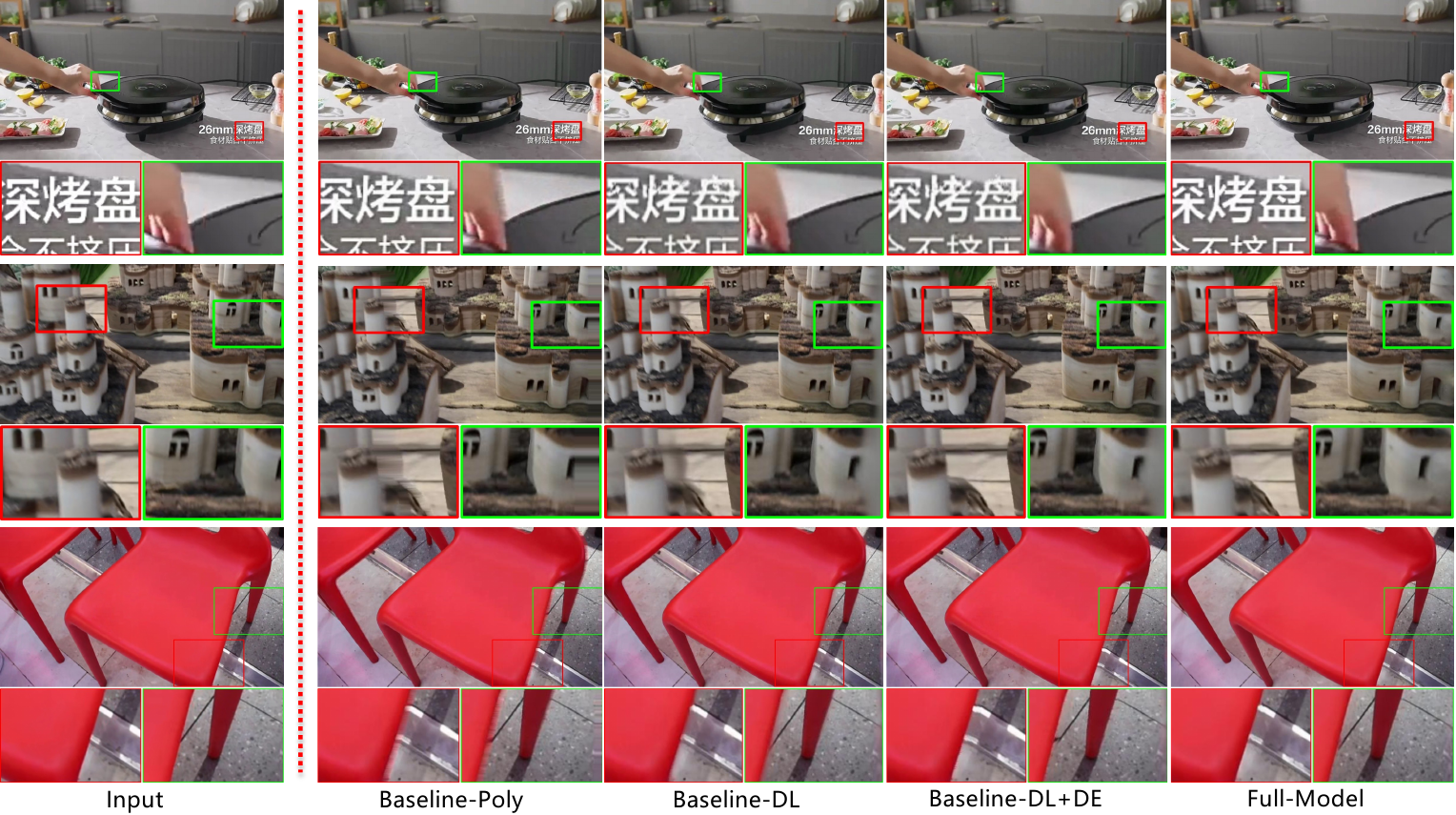}
        \caption{Ablation study of our model. Our full model can effectively integrate and optimize the results of different strategies, resulting in more realistic and natural outputs.}
        \label{fig_ablation}
        \vspace{-15pt}
    \end{figure}
    
    To evaluate these variants, we use the same metrics mentioned above. Fig.~\ref{fig_ablation} presents the qualitative results, while the quantitative results are shown in Table~\ref{tab:ablation}.
    
    \vspace{2pt}\noindent\textbf{Efficacy of Disparity Expansion Strategy.}
    We compare the `Baseline-DL'  with `Baseline-DL + DE' to evaluate the efficacy of our disparity expansion strategy. As shown in Fig.~\ref{fig_DE} and second row of Fig.~\ref{fig_ablation}, 
    `Baseline-DL' tends to generates incorrect foreground content near the foreground-background edges, resulting in foreground bleeding. This occurs because it relies on surrounding pixels for content filling, but warped edges lack reference background information, leaving only foreground context. By contrast, `Baseline-DL+DE' employs our disparity expansion strategy, which supplies additional background information at these edges, preventing foreground bleeding and enabling accurate and natural background content filling. Table~\ref{tab:ablation} further shows that `Baseline-DL + DE' outperforms `Baseline-DL' across all four metrics, enhancing overall generation quality.

    \vspace{2pt}\noindent\textbf{Efficacy of MHFU Refiner.}
    Our MHFU refiner synthesizes the final output by integrating and refining results from the multi-branch inpainting modules. To evaluate its effectiveness, we compare the `Full-Model' with `Baseline-Poly', `Baseline-DL', and `Baseline-DL+DE'. 
    As shown in Fig.~\ref{fig_ablation}, `Baseline-Poly' maintains text stability but suffers from pixel stretching in edge areas, leading to artifacts. `Baseline-DL' generates reasonable content and preserves geometric properties, but lacks background information at foreground-background edges, causing foreground bleeding. `Baseline-DL+DE' employs disparity expansion to supply background information, guiding correct background filling, yet it may introduce geometric distortions in complex regions, such as converting straight lines into polylines. Additionally, all branches struggle with fine content generation at the far right image edge.
    The MHFU refiner addresses these issues by fusing and refining the results, achieving a realistic and natural output that balances text stability, accurate background filling, geometric integrity, and fine content at the far right edge of image. Quantitative results in Table~\ref{tab:ablation} confirm that the `Full-Model' with the MHFU refiner significantly improves performance across all four metrics, showing its superiority.

    \begin{figure}[!t]
        \centering
        \includegraphics[width=3.35in, angle=0]{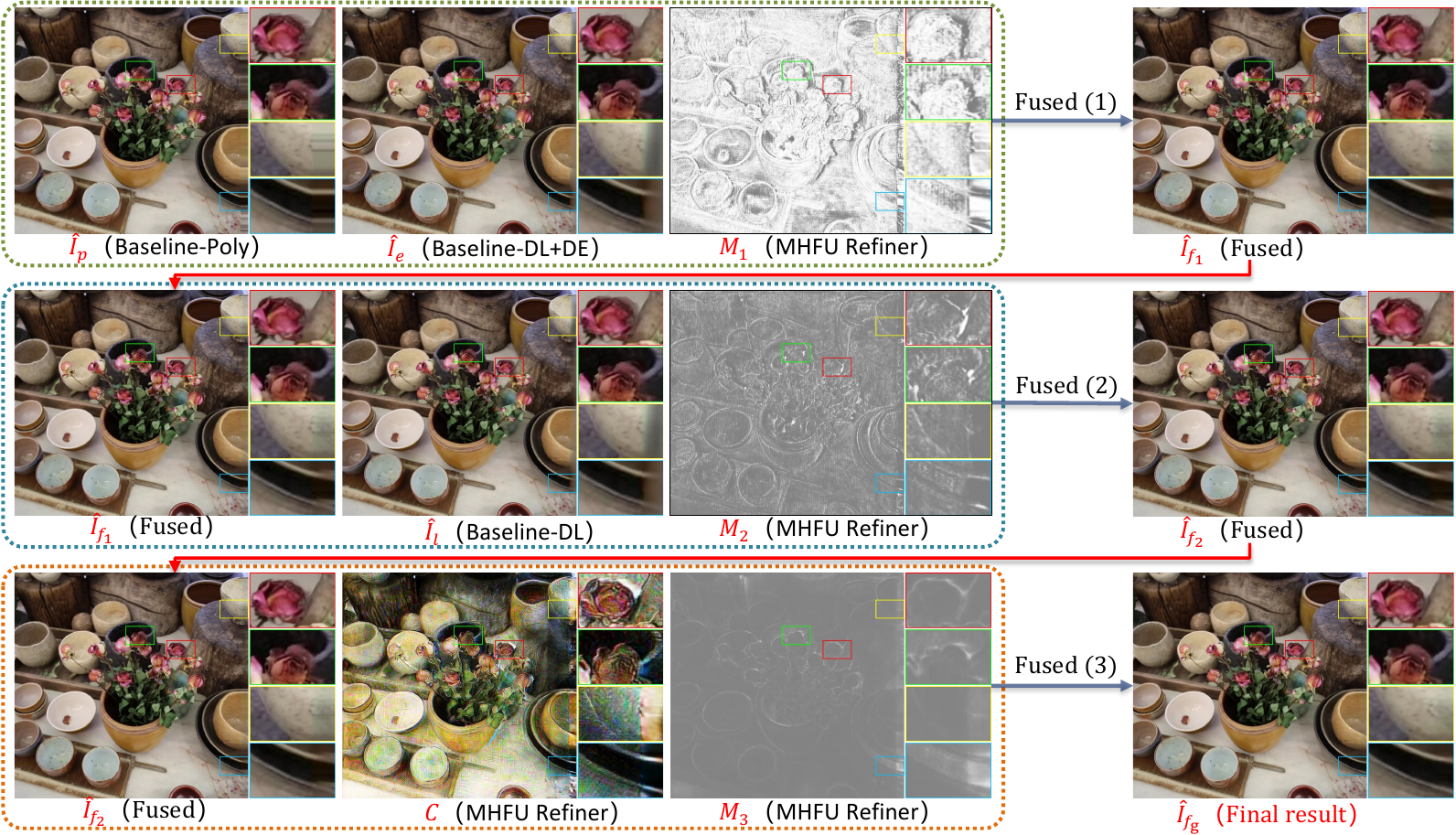}
        \caption{Visualization of our MHFU refiner. By generating ${M}_{1}$, ${M}_{2}$, ${M}_{3}$, and $C$, as well as using them to fuse the $\hat{I}_{p}$, $\hat{I}_{e}$, and $\hat{I}_{l}$ through the Equations \ref{f1}, \ref{f2}, and \ref{f3}, the MHFU refiner achieves more realistic results with refined local details.}
        \vspace{-13pt}
        \label{fig_MHFU}
    \end{figure}

    To further demonstrate the effectiveness of our MHFU refiner, we visualize the results of each stage in Fig.~\ref{fig_MHFU}. The generated masks highlight a focus on edge areas, which are most prone to needing filling. Using ${M}_{1}$, ${M}_{2}$, and Equations \ref{f1} and \ref{f2}, we integrate advantageous content from $\hat{I}_{p}$, $\hat{I}_{e}$, and $\hat{I}_{l}$ to obtain $\hat{I}_{f_2}$. Despite this integration, some incorrect patterns and artifacts appear at certain edges, such as the edges of flowers and the far right of image. The MHFU refiner then focuses on these areas and uses ${M}_{3}$, content map $C$, and Equation \ref{f3} for further refinement, resulting in the final output.

    Overall, our `Full-Model' achieves the best results for both quantitative and qualitative evaluations, demonstrating the effectiveness of each module in our approach.

    \vspace{-2pt}\section{Conclusion}\label{sec:conclution}
    In this paper, we propose SpatialMe, a novel stereo video conversion framework based on depth-warping and blend-inpainting. To address the shortage of data, we introduce StereoV1K, a high-quality real-world stereo video dataset for benchmarking stereo video generation, which can significantly advance progress in this area. Additionally, we present a MHFU refiner, which integrates and optimizes the outputs from our multi-branch inpainting module. We also propose a disparity expansion strategy to overcome the problem of foreground bleeding. Extensive experiments justify that our approach enables high-fidelity stereo video conversion, outperforming previous state-of-the-art methods.


\bibliographystyle{IEEEbib}

\end{document}